\documentclass{article}

\PassOptionsToPackage{numbers}{natbib}

\usepackage[final]{nips_2018}

\usepackage[utf8]{inputenc} 
\usepackage[T1]{fontenc}    
\usepackage{hyperref}       
\usepackage{url}            
\usepackage{booktabs}       
\usepackage{amsfonts}       
\usepackage{nicefrac}       
\usepackage{microtype}      
\usepackage{times}
\usepackage{helvet}
\usepackage{courier}
\usepackage{url}
\usepackage{graphicx}
\usepackage{amsthm}
\usepackage{amsmath}
\usepackage{epic}
\numberwithin{equation}{section}
\usepackage{dsfont}
\usepackage{algorithm}
\usepackage[noend]{algpseudocode}
\usepackage{amsbsy}
\usepackage{amssymb}
\usepackage{nameref}
\usepackage{tikz}
\usetikzlibrary{fit,positioning,arrows,automata}
\DeclareMathOperator*{\argmax}{argmax} 

\title{Patient Subtyping with Disease Progression and Irregular Observation Trajectories}

\author{
  Nikhil Galagali and Minnan Xu-Wilson \\
  Philips Research North America \\
  nikhilg18@gmail.com, minnan.xu@philips.com
}

\begin{document}
\maketitle
\begin{abstract}
Patient subtyping based on temporal observations can lead to significantly nuanced subtyping that acknowledges the dynamic characteristics of diseases. Existing methods for subtyping trajectories treat the evolution of clinical observations as a homogeneous process or employ data available at regular intervals. In reality, diseases may have transient underlying states and a state-dependent observation pattern. In our paper, we present an approach to subtype irregular patient data while acknowledging the underlying progression of disease states. Our approach consists of two components: a probabilistic model to determine the likelihood of a patient's observation trajectory and a mixture model to measure similarity between asynchronous patient trajectories. We demonstrate our model by discovering subtypes of progression to hemodynamic instability (requiring cardiovascular intervention) in a patient cohort from a multi-institution ICU dataset. We find three primary patterns: two of which show classic signs of decompensation (rising heart rate with dropping blood pressure), with one of these showing a faster course of decompensation than the other.  The third pattern has transient period of low heart rate and blood pressure. We also show that our model results in a 13\% reduction in average cross-entropy error compared to a model with no state progression when forecasting vital signs.
\end{abstract}

\section{Introduction}\label{sec:introduction}
\noindent
Patient subtyping is an important topic in medical informatics. Subtyping can be used to make improved predictions, understand disease etiologies and healthcare practices, plan customized treatments, and design efficient clinical trials \citep{Marlin2012,Chang2011,Gundlapalli2008,Kohane2011}. The medical community has begun to recognize that many diseases have heterogeneous underlying mechanisms and phenotypes \citep{DeKeulenaer2009,Suratt2017}. With increasing amounts of patient data being collected in electronic medical records and large medical databases, there is a great opportunity to mine data and identify patient subtypes to better patient care and improve outcomes. Patient subtyping can also have a direct economic impact on the healthcare system by characterizing current patient-provider interactions and suggesting optimal resource management.  Today's medical databases have patients' longitudinal trajectories of observations and interventions charted along with their time stamps. Subtyping patients based on their temporal observation trajectories could lead to significantly nuanced subtyping that acknowledges the dynamic characteristics of diseases. Traditionally, though, patient subtyping has been based either on aggregates of patient's entire time course of observations or summaries over blocks of fixed time intervals. These approaches result in vectors of fixed size for all patients, which are then amenable to well-known clustering approaches such as K-means clustering, hierarchichal clustering etc \citep{Vranas2017,Marlin2012,Cohen2010}. Native data in medical records, however, are almost always incomplete and irregularly observed over varying time intervals. The missingness of data is typically tackled by imputation to produce a complete dataset of constant dimension. Although practically useful, clustering with summary-based fixed dimensional dataset ignores the rich information in the temporal patterns of clinical observations/interventions. Methods that subtype raw observation trajectories are being developed in recent times \citep{Doshi-Velez2014,Schulam2015,Saria2010}. Existing methods, however, either treat the evolution of clinical markers as a homogeneous process or employ data available at regular intervals; in reality diseases may have transient underlying states and a state-dependent observation pattern.
%
In this paper, we present an approach to subtype raw unsummarized data from electronic medical records, while acknowledging the underlying progression of disease states. Our approach consists of two components: a probabilistic model to determine the likelihood of a patient's observation trajectory and a mixture model to measure similarity between asynchronous patient trajectories. 
\section{Methodology}\label{sec:methodology}
\noindent
Consider that we have data from $N$ patients, each associated with their time course of observations given by ${Y_n} \equiv \{Y_{n,t_{1}},Y_{n,t_{2}}, ..., Y_{n,t_{n}}\}$. Here, $Y_{n,t_{i}}$ is the vector of observations at time $t_i$ and ${Y_n}$ is the trajectory of observation vectors of patient $n$. The length of each $Y_{n,t_i}$ is $D$, where $D$ is the number of features that could be observed. For example, if we have the heart rate, blood pressure, and respiratory rate measurements of patients, $D$ would be three. Usually, only a subset of the $D$ features are actually observed at any time, with the specific features observed being different at each time point. This results in an incomplete observation set with many feature observations missing. As such, patient observations are made when appropriate---when the patient appears for a routine check-up or when clinicians ask for specific tests/measurements. As a result, the trajectories of patient observations do not synchronize in time or the type of features that are observed. 
\subsection{Patient disease trajectory model}\label{sec:patientdiseasetrajectory}
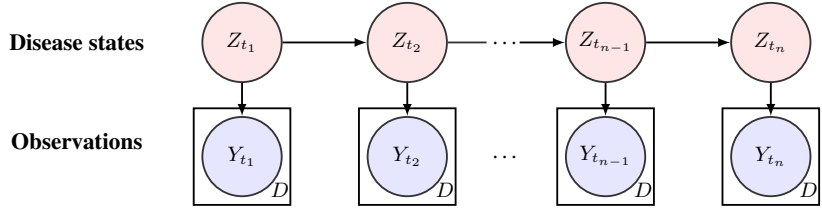
\begin{figure*}
\centering
\resizebox{0.8\textwidth}{!}{

\begin{tikzpicture}
\tikzstyle{main}=[circle, minimum size = 12mm, thick, draw =black!80, node distance = 5mm]
\tikzset{box/.style={draw, minimum size=2em, text width=4.5em, text centered},
         bigbox/.style={draw, thick, minimum size=1.2cm,minimum height=1.2cm,label={[align=right,shift={(-1.0ex,3ex)}]south east:\llap{#1}}}
}
\tikzstyle{connect}=[-latex, thick]
\tikzstyle{box}=[rectangle, draw=black!100]
  \node[box,draw=white!100] (Latent) at (-1.5,0) {\textbf{Disease states}};
  \node[main,fill=red!10] (Zt1) at (1,0) {\small $Z_{t_1}$};
  \node[main,fill=red!10] (Zt2) at (3.5,0) {\small $Z_{t_2}$};
  \node[main,fill=red!10] (ZtN1) at (6.5,0.0) {\small $Z_{t_{n-1}}$};
  \node[main,fill=red!10] (ZtN) at (9.0,0.0) {\small $Z_{t_{n}}$};
  \node[box,draw=white!100] (Observed) at  (-1.5,-1.5){\textbf{Observations}};
  \node[main,fill=blue!10] (Yt1) [right=of Observed,below=of Zt1] {\small $Y_{t_1}$};
  \node[main,fill=blue!10] (Yt2) [right=of Yt1,below=of Zt2] {\small $Y_{t_2}$};
  \node[main,fill=blue!10] (YtN1) [right=of Yt2,below=of ZtN1] {\small $Y_{t_{n-1}}$};
  \node[main,fill=blue!10] (YtN) [right=of YtN1,below=of ZtN] {\small $Y_{t_{n}}$};
  \node[bigbox={\footnotesize $D$},label= south east:, fit=(Yt1)] (E) {};
  \node[bigbox={\footnotesize $D$},label= south east:, fit=(Yt2)] (F) {};
  \node[bigbox={\footnotesize $D$},label= south east:, fit=(YtN1)] (F) {};
  \node[bigbox={\footnotesize $D$},label=south east:, fit=(YtN)] (D) {};
  \path (Zt2) -- node[auto=false]{\ldots} (ZtN1);
  \path (Zt1) edge [connect] (Zt2)
        (ZtN1) edge [connect] (ZtN)        
        (Yt2) -- node[auto=false]{\ldots} (YtN1);
  \path (Zt1) edge [connect] (Yt1);
  \path (Zt2) edge [connect] (Yt2);
  
  \path (ZtN1) edge [connect] (YtN1);  
  \path (ZtN) edge [connect] (YtN);
  \draw[dashed]  [below=of Zt1,above=of Yt1];
  \draw [-,thick,black!80](Zt2.east)to[out=0,in=180] node[above]{}(4.7,0);
  \path (5.2,0) edge [connect] (ZtN1);
\end{tikzpicture}}
\caption{Continuous time HMM describing patients' disease and observation progression}
\label{CTMC}
\end{figure*}

\noindent 
Disease evolution is fundamentally a continuous process: patient's disease state transitions can happen at any time with the chance of state transition between any two time points higher if the time interval is longer. We thus use a continuous-time Markov chain to model the evolution of a patient's disease state (See Appendix for details). The disease state of patient $n$ at time $t_i$ is denoted as $Z_{n,t_{i}}$ and takes one of a set of discrete values. The disease state is naturally hidden, i.e., we never get to observe the actual disease state. In fact, the precise definition of the disease states is apriori unknown. The disease states can be learnt from data in an unsupervised manner and subsequently the states interpreted based on the parameters that describe the states. The observation vectors $Y_{n,t_{i}}$ are surrogates of the underlying disease state $Z_{n,t_{i}}$. To reflect this behaviour in our model, we model the observations by a conditionally independent probability model $P(Y_{n,t_{i}} \vert Z_{n,t_i})$, where observation $Y_{n,t_{i}}$ is independent of all other observations $Y_{n,t_{j}}$ given the current disease state $Z_{n,t_{i}}$. Overall, our model for the patient's observation trajectory can be described by the continuous-time hidden Markov model (CT-HMM) shown in Figure \ref{CTMC}.
%
The probability of the trajectory of patient $n$ is given by
{\footnotesize
\begin{equation}
P(Z_{n,t_1:t_n},Y_{n,t_1:t_n})=P(Y_{n,t_1:t_n} \vert Z_{n,t_1:t_n} )P(Z_{n,t_1:t_n})
\label{jointDiseaseObservations}
\end{equation} }
\noindent
We use the notation \emph{l:r} to denote all values ranging from $l$ to $r$ (inclusive of both boundaries). 
A common modeling choice that we incorporate in our model is that the features are conditionally independent given the corresponding disease state, i.e.,
{\footnotesize
\begin{equation}
P(Y_{n,t_i} \vert Z_{n,t_i})=\prod_{d=1}^{D}P(Y_{n,t_i,d} \vert Z_{n,t_i})
\end{equation}}
\noindent
The choice of the conditional distribution of the observation $Y_{n,t_i,d}$ given the disease state $Z_{n,t_i}$ can be made as per context. 
Specifically, if we assume that a feature $d$ can fall into one of $J$ bins, the conditional probability of the $j$th bin is given by
{\footnotesize
\begin{equation}
P(Y_{n,t_i,d}=j \vert Z_{n,t_i}=k)=w_{k,d,j},
\end{equation}}
\noindent
where $k$ refers to the disease state at time $t_i$ and $w_{k,d,1:J}$ are the parameters of the categorical distribution of feature $d$ given disease state $k$. By construction $\sum_{j}w_{k,d,j}=1$.\ In case of missing feature observation, we marginalize that observation from the model.
The disease states in a patient's observation timeline are unknown. Thus, we can quantify the patient's observation trajectory by marginalizing the disease state out of Equation \ref{jointDiseaseObservations}, giving
{\footnotesize
\begin{equation}
P(Y_{n,t_1:t_n})=\sum_{Z_{n,t_1:t_n}}P(Z_{n,t_1:t_n},Y_{n,t_1:t_n}).
\label{marginalizedDiseaseTrajectory}
\end{equation}}
\noindent
This is we refer to as the likelihood under the patient's \emph{disease trajectory model}. 
Equipped with a likelihood model of the patient's observation trajectory, we are now in a position to describe a measure of similarity between trajectories of different patients. Patients whose observation trajectories are more probable under a disease trajectory model than other trajectory models can be considered to be similar trajectories. 
%
\subsection{Mixture model}
\noindent
We subtype patients into different clusters using the mixture model. Consider that we are interested in identifying $M$ subtypes among the patients. 
The joint distribution of patient $n$'s subtype assignment $m_{n}$ and his/her trajectory $Y_{n,t_1:t_n}$ is given by
{\footnotesize
\begin{equation}
P(m_{n},Y_{n,t_1:t_n})=P(m_n)P(Y_{n,t_1:t_n} \vert m_{n}).
\end{equation}}
\noindent
Here $m_{n} \in \{1,2,...,M\}$ and $P(Y_{n,t_1:t_n} \vert m_{n})$ is evaluated using the disease trajectory model corresponding to subtype $m_n$. To infer the subtypes, we identify patient subtype assignments and subtype parameters so as to maximize the joint probability of the subtype assignment and the conditional observation trajectory probability over all patients. Mathematically, assuming independence of patients, the objective used to identify the subtypes is
{\footnotesize
\begin{align}
\{{m^{*}_{1:N}},{\pi^{*}_{1:M}},{Q^{*}_{1:M}},{w^{*}_{1:M}} \} =\argmax\limits_{{m_{1:N},\pi_{1:M},Q_{1:M},w_{1:M}}} \prod_{n=1}^{N} P(m_{n},Y_{n,t_1:t_n}),
\label{hardEMobjective}
\end{align}}
\noindent
where $\pi_{1:M},Q_{1:M},w_{1:M}$ are subtype-specific parameters. With the above objective, each patient gets assigned the subtype with the highest posterior probability. 
Once the optimal parameters are learnt from the training data, for a new patient not in the training data, the subtype is identified as the subtype with the highest posterior probability for that patient. Complete details of steps involved in subtype learning are in the Appendix.
\section{Results}
\subsection{Data summary}
\noindent
We applied our subtyping model to a cohort of 6972 hemodynamically unstable patients obtained from the eResearch Institute \citep{McShea2010}. 
For our subtyping analyses, we only consider heart rate and noninvasive systolic blood pressure (nSBP) measurements among various observations made for the patients. Interventions were given after all heart rate and blood pressure observations. The allowed range of heart rate values we consider is 40--150 beats/min and that of systolic blood pressure is 40--200 mmHg. All outliers are treated as missing observations. The observations for both features are discretized into five equal-sized bins. We tried a few different bin sizes and found the results to be similar. 
%
%
\subsection{Inferring subtypes of hemodynamically unstable patients}
\begin{figure}
\centering
\includegraphics[width=0.85\textwidth]{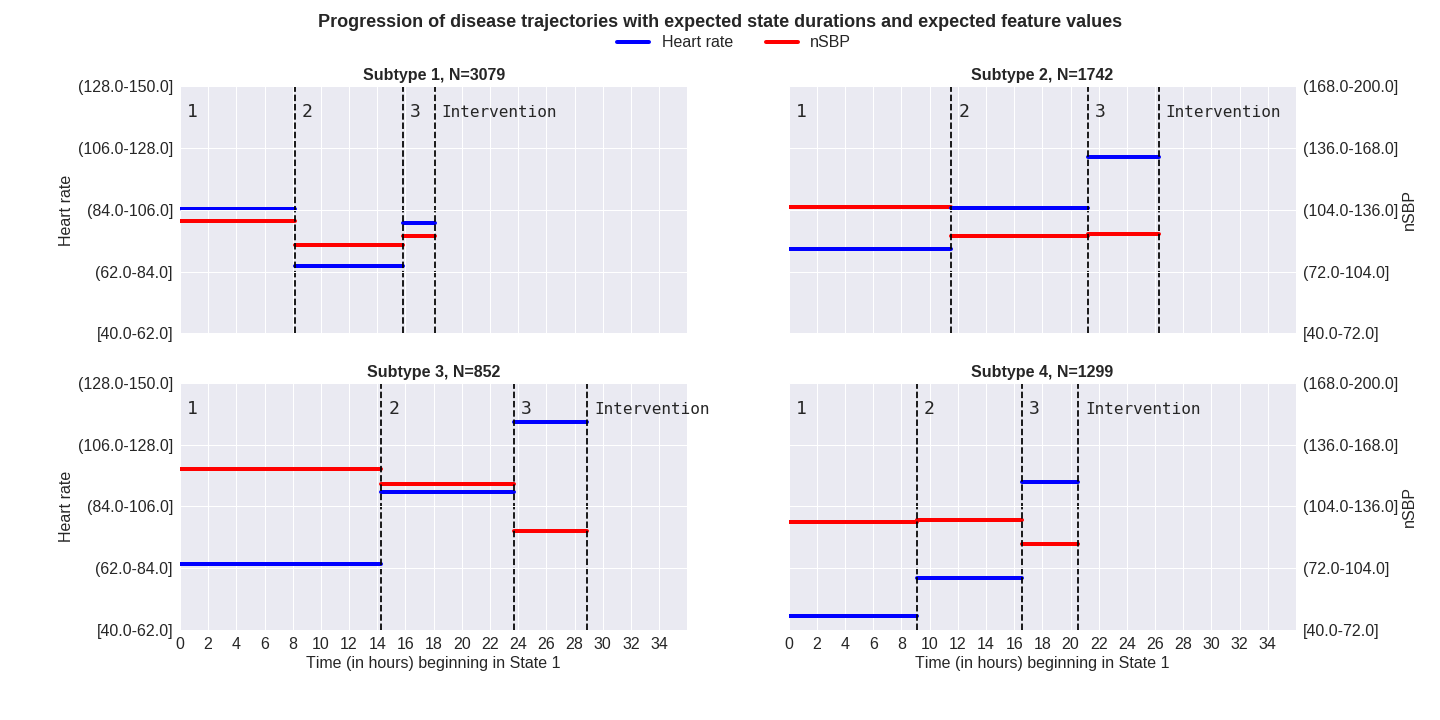}
\caption{Progression trajectory of patients starting in disease state 1 for the four subtypes}
\label{expectedDiseaseTrajectory}
\end{figure}

\noindent
We use our subtyping algorithm to identify 4 clusters using the time course of heart rate and blood pressure measurements and also the intervention time. Thus, effectively, we have three features in the analysis: heart rate, systolic blood pressure, and an indicator of treatment administration. 
We force the administration of an intervention to be the indicator of the final disease state in our model. 
 As a result, all prior disease states can be viewed relative to the final intervention state. 
We set the number of allowed disease states to be 4 and impose a uniform prior distribution on the subtype of all patients. We allow the initial probability vector to be nonzero for all four disease states. Further, for the present analysis, we only allow state transitions to the next state, although this assumption can be relaxed in general. 
The progression of disease trajectories assuming the patients arrive in state 1 of the inferred subtypes are shown in Figure \ref{expectedDiseaseTrajectory}. To compute the progression trajectory, we calculate the expected duration of the four disease states and compute the expected values of the features in each disease state. 
We find three primary patterns: subtype 2, 3, and 4 show classic signs of decompensation (rising heart rate with dropping blood pressure), with subtype 4 showing a faster course of decompensation than the other two. Subtype 1 has transient period of low heart rate and blood pressure. In general, across all subtypes we also observe that state 3 has the shortest duration among all disease states. This corroborates a common observation of quick deterioration---a period of acute changes over a short time---before intervention among ICU patients.
\subsection{Model evaluation by prediction}
\noindent
We evaluate our subtyping algorithm through quantitative tests of prediction on future observations. For this we split the above cohort into training and test sets in a 80:20 proportion. We only consider the vital signs (heart rate and blood pressure values) for this analysis and do not include the indicators of intervention. The evaluation of prediction accuracy is performed as follows. First we use the training data to learn the subtypes. Once again a uniform prior distribution is imposed on the subtype of each patient. Having learnt the underlying subtype parameters, for each patient in the test data, we use the first 70\% of the timepoints when observations are made to identify the patient's subtype. Knowing the patient's subtype, we predict the observations at the remaining 30\% of the timepoints. We compare the predicted heart rate and blood pressure bin probabilities with the available data points to compute the average cross-entropy error over the latter 30\% observations.  The total forecasting error is obtained by computing the mean of the cross-entropy error over all patients in the test set. The estimates of forecasting error and their standard errors with different number of subtypes and disease states are given in Table \ref{predictionAccuracy}.
\begin{table}[h]
\centering
{\scriptsize
\begin{tabular}{|ccc|ccc|ccc|ccc|ccc|ccc|ccc|}
\hline
\multicolumn{1}{|c|}{Subtypes} &\multicolumn{6}{c|}{Disease states} \\
\hline
\multicolumn{1}{|c|}{} &\multicolumn{1}{c|}{1} &\multicolumn{1}{c|}{2} &\multicolumn{1}{c|}{3} &\multicolumn{1}{c|}{4} &\multicolumn{1}{c|}{5} & \multicolumn{1}{c|}{6}  \\
\hline
\multicolumn{1}{|c|}{1} & \multicolumn{1}{c|}{$1.39 \pm 0.015$} & \multicolumn{1}{c|}{$1.09 \pm 0.016 $} & \multicolumn{1}{c|}{$1.01 \pm 0.017 $} & \multicolumn{1}{c|}{$0.91 \pm 0.018 $} & \multicolumn{1}{c|}{$0.90 \pm 0.017 $} & \multicolumn{1}{c|}{$0.89 \pm 0.019 $} \\
\multicolumn{1}{|c|}{2} & \multicolumn{1}{c|}{$1.11 \pm 0.018 $} & \multicolumn{1}{c|}{$0.96 \pm 0.018 $} & \multicolumn{1}{c|}{$0.87 \pm 0.018 $} & \multicolumn{1}{c|}{$0.87 \pm 0.019 $} & \multicolumn{1}{c|}{$0.84 \pm 0.019 $} & \multicolumn{1}{c|}{$0.84 \pm 0.019 $} \\
\multicolumn{1}{|c|}{3} & \multicolumn{1}{c|}{$1.00 \pm 0.018 $} & \multicolumn{1}{c|}{$0.90 \pm 0.018 $} & \multicolumn{1}{c|}{$0.84 \pm 0.018 $} & \multicolumn{1}{c|}{$0.83 \pm 0.019 $} & \multicolumn{1}{c|}{${\bf 0.79 \pm 0.019} $} & \multicolumn{1}{c|}{$0.81 \pm 0.018 $} \\
\multicolumn{1}{|c|}{4} & \multicolumn{1}{c|}{$0.92 \pm 0.018 $} & \multicolumn{1}{c|}{$0.87 \pm 0.018 $} & \multicolumn{1}{c|}{$0.82 \pm 0.018 $} & \multicolumn{1}{c|}{$0.80 \pm 0.019 $} & \multicolumn{1}{c|}{$0.81 \pm 0.019 $} & \multicolumn{1}{c|}{$0.82 \pm 0.02 $} \\
\multicolumn{1}{|c|}{5} & \multicolumn{1}{c|}{$0.89 \pm 0.017 $} & \multicolumn{1}{c|}{$0.84 \pm 0.018 $} & \multicolumn{1}{c|}{$0.82 \pm 0.019 $} & \multicolumn{1}{c|}{$0.81 \pm 0.019 $} & \multicolumn{1}{c|}{$ - $} & \multicolumn{1}{c|}{$ - $} \\
\multicolumn{1}{|c|}{6} & \multicolumn{1}{c|}{${\bf 0.88 \pm 0.019} $} & \multicolumn{1}{c|}{$0.85 \pm 0.02 $} & \multicolumn{1}{c|}{$ -  $} & \multicolumn{1}{c|}{$ - $} & \multicolumn{1}{c|}{$ - $} & \multicolumn{1}{c|}{$ - $} \\
\hline 
\end{tabular}}
\caption{\footnotesize Forecasting cross-entropy error with different number of subtypes and disease states} 
\label{predictionAccuracy}
\end{table}
%
We see that there is a consistent reduction in the error with the increase in disease states for all subtype numbers. In particular, there is a $13\%$ reduction in forecasting error in predicting patient trajectories with optimal number of subtypes with multiple disease states as compared to optimal subtyping model with only one disease state. These results demonstrate the potential of improved subtyping and more accurate forecasting by subtyping patient disease trajectories while acknowledging the underlying disease progression. A natural direction of future work would be to learn subtypes of observation trajectories with a richer class of state-conditioned observation models, e.g., models with polynomial basis.

\clearpage
\newpage
\newpage

\bibliographystyle{plain}
\bibliography{galagali}

\clearpage
\newpage
\newpage

\section{Appendix}

\subsection{Continuous time Markov chains}
\noindent
As mentioned in the main text, we model the evolution of disease state $P(Z_{n,t_i} \vert Z_{n,t_{i-1}})$ with a continuous-time Markov chain. A continuous-time Markov chain is a continuous-time process on a state-space (here the different disease states) satisfying the Markov property. This means that if $\mathcal{F}_{Z(s)}$ is all the information about the history of the disease state $Z$ up to time $s$ and $s \leq t$, then Z(t) is independent of all $Z(t')$, where $t'<s$, given Z(s). Mathematically, this can be expressed as
{\footnotesize
\begin{equation}
P(Z(t)=k \vert \mathcal{F}_{Z(s)})=P(Z(t)=k \vert Z(s)).
\label{eq1}
\end{equation}}
\noindent
Further, we assume the process to be time-homogeneous, so that
{\footnotesize
\begin{equation}
P(Z(t)=k \vert Z(s))=P(Z(t-s)=k \vert Z(0)).
\label{eq2}
\end{equation}}
\noindent
Equations \ref{eq1} and \ref{eq2} define a time-homogeneous continuous-time Markov chain and model the disease state evolution in our model. We allow $K$ different disease states in our model. The transition probability of moving from state $a$ to state $b$ over time $\Delta$ in a continuous-time Markov chain is given by 
{\footnotesize
\begin{align}
&P(Z_{n,t_i}=b \vert Z_{n,t_{i-1}}=a,t_{i}-t_{i-1}=\Delta;Q) \nonumber \\
              &=expm(\Delta Q)_{ab},
\label{transitionProb}              
\end{align}}
\noindent
where $Q$ is the generator matrix of the Markov process and $expm$ is the matrix exponential. The probability of the initial state $P(Z_{n,t_1})$ is parameterized by $\pi=\{\pi_{1},\pi_{2},...,\pi_{K}\}$ and given by
{\footnotesize
\begin{equation}
\pi_{k} \triangleq P(Z_{n,t_1}=k),\mbox{  }k=1,2,...,K
\end{equation}}
\begin{algorithm}[h]
\begin{algorithmic}[1]
\State \textbf{Given}: Number of subtypes $M$ and observation trajectories \{${Y_n} \equiv Y_{n,t_1:t_n}$\} of N patients
\newline
\State {\bf Repeat until convergence:}
\newline
\State\hspace{0.5cm} {\bf Step 1}:
{\footnotesize
\begin{equation}
{m^{*}_{1:n}}=\argmax_{m_{1:n}} P({\bf Y_{1:n}},{m_{1:n}},\bar{\pi},\bar{Q},\bar{w})
\label{estep}
\end{equation}}
\State\hspace{0.5cm}{\bf Step 2}:
{\footnotesize
\begin{align}
&\mbox{For m=1 to M:}\nonumber \\
&\{\pi_{m}^{*},Q_{m}^{*},w_{m}^{*}\}=\nonumber \\ &\argmax\limits_{{\pi_{m},Q_{m},w_{m}}} \prod_{n \in N(m)} P(Y_{n,t_1:t_i};{\pi_{m},Q_{m},w_{m}}),\nonumber \\ &\mbox{ where }N(m)\mbox{ are patients assigned to subtype }m
\label{mstep}
\end{align}}
\newline
\end{algorithmic}
\caption{Patient subtyping algorithm}
\label{alg:patientsubtyping}
\end{algorithm}
\subsection{Subtype learning}
\noindent
We now present the different steps involved in training the subtyping model. We train our subtyping model by maximizing the objective (Equation \ref{hardEMobjective}) using a coordinate ascent optimization algorithm. The algorithm consists of two alternating steps: Step 1, when each patient trajectory gets assigned to the subtype with the highest posterior probability, and Step 2, when all patients assigned to a subtype are used to optimize the parameters of that subtype. The precise mathematical forms of the two steps are given in Algorithm \ref{alg:patientsubtyping}. In Step 2 of the above algorithm, parameters of each subtype are learnt by training the disease trajectory model described in Section \ref{sec:patientdiseasetrajectory}. The solution of each maximization problem in Step 2 is a maximum likelihood estimate of the subtype parameters with the data assigned to that subtype. As was explained in Section \ref{sec:patientdiseasetrajectory}, the likelihood of a patient trajectory $P(Y_{n,t_1:t_n})$ can be realized by marginalizing the hidden disease states from the joint probability distribution of the observations and the disease states (Equation \ref{marginalizedDiseaseTrajectory}). Thus, the optimization in Step 2 can be solved with the expectation maximization algorithm. The E- and M- steps in optimizing equation \ref{mstep} are given in Algorithm \ref{alg:solutiondiseaseprogression}.

\begin{algorithm}[h]
\begin{algorithmic}[1]
\State \textbf{Given}: Trajectories ${\bf Y} \equiv\{{{Y}_{n \in N(m)}}$\} of patients assigned to subtype $m$
\newline
\State ${Z_{n}} \equiv Z_{n,t_1:t_n}$ and ${{\bf Z} \equiv \{Z_{n \in N(m)}\}}$
\newline
\State {\bf Repeat until convergence}
\newline
\State\hspace{0.5cm} {\bf E-Step}:
{\footnotesize
\begin{align}
&\mathbb{E}_{P({\bf Z},{\bf Z}(t) \vert {\bf Y}; \pi',Q',w')} \log P({\bf Y},{\bf Z},{\bf Z}(t);\pi,Q,w) \nonumber\\ &=\mathbb{E}_{P({\bf Z},{\bf Z}(t) \vert {\bf Y};\pi',Q',w')} \log P({\bf Z},{\bf Z}(t);Q) \nonumber \\ &+\mathbb{E}_{P({\bf Z} \vert {\bf Y};\pi',Q',w')} \log P({\bf Y} \vert {\bf Z};w)
\label{estepdiseaseProgression}
\end{align}}
\State\hspace{0.5cm}{\bf M-Step}:
{\footnotesize
\begin{align}
&\pi_{m},Q_{m},w_{m}=\nonumber \\ &\argmax_{\pi,Q,w} {\mathbb{E}_{P({\bf Z},{\bf Z}(t) \vert {\bf Y})} \log P({\bf Y},{\bf Z},{\bf Z}(t);\pi,Q,w)}
\label{mstep_diseaseProgression}
\end{align}}
\end{algorithmic}
\caption{Disease trajectory learning}
\label{alg:solutiondiseaseprogression}
\end{algorithm}
\noindent
The E-Step and M-Step in Algorithm \ref{alg:solutiondiseaseprogression} can be simplified for our construction of the patient disease trajectory model. The expectation of the first term of the RHS in Equation \ref{estepdiseaseProgression} is given by
{\footnotesize
\begin{align}
&\mathbb{E}_{P({\bf Z},{\bf Z}(t) \vert {\bf Y}; \pi',Q',w')}\log P({\bf Z},{\bf Z}(t);\pi,Q)\nonumber \\
&=\sum_{\Delta}\sum_{a,b \in K}C_{ab}(\Delta)(\sum_{c,d \in [K]}(\log Q_{cd})\mathbb{E}[\mathcal{N}_{cd}(\Delta) \vert {\bf Z};Q']\nonumber \\
&\hspace{1cm}-Q_{cd}\mathbb{E}[R_{c}(\Delta) \vert {\bf Z};Q'])+\nonumber\\
&\hspace{1cm}\mathbb{E}_{P({\bf Z} \vert {\bf Y}; \pi',Q',w')}\log P({\bf Z}_{t_1};\pi),
\end{align}}
\noindent
where
{\footnotesize
\begin{align}
&C_{ab}(\Delta) \triangleq \nonumber \\ &\sum_{n}\sum_{t_2}^{t_{n}}P(Z_{n,t_{i-1}}=a,Z_{n,t_{i}}=b \vert {\bf Y};\pi',Q')\mathds{1}_{t_{i}-t_{i-1}=\Delta},
\end{align}}
$\mathcal{N}_{ab}(\Delta)$ is the number of transitions between states $a$ and $b$ in time $\Delta$ and $\mathcal{R}_{a}(\Delta)$ is the duration of time spent in state $a$ during time interval $\Delta$. A detailed derivation of the above expression can be found in \cite{Metzner2007}. The second term of the RHS in Equation \ref{estepdiseaseProgression} can be written as
{\footnotesize
\begin{align}
&\mathds{E}_{P({\bf Z},\vert {\bf Y}; \pi',Q',w')} \log P({\bf Y} \vert {\bf Z})\nonumber \\
                                                                                                                  &=\sum_{n}\sum_{t=t_1}^{t_n}\sum_{k=1}^{K}\sum_{d=1}^{D} \gamma_{n,t,k} P(Y_{n,t,d} \vert Z_{n,t}=k) \nonumber\\
                                                                                                                  &=\sum_{n}\sum_{t=t_1}^{t_n}\sum_{k=1}^{K}\sum_{d=1}^{D} \gamma_{n,t,k} \sum_{j} w_{k,d,j}^{\mathds{1}_{Y_{n,t,d}=j}\mathds{1}_{Y_{n,t,d}}},
\label{observation_expectation}                                                                                                                  
\end{align}}
\noindent
where $\gamma_{n,t,k}=P(Z_{n,t}=k \vert Y_n)$ is the posterior probability of disease state $k$ for patient $n$ at time point $t$, 
 $\mathds{1}_{Y_{n,t,d}}$ is an indicator of feature $d$ not missing at time point $t$ and $\mathds{1}_{Y_{n,t,d}=j}$ is an indicator function for observation $Y_{n,t,d}$ belonging to the $j^{th}$ discrete bin.

The M-step in Equation \ref{mstep_diseaseProgression} results in the following closed-form expressions for the parameters of the observation model and the initial probability vector:
{\footnotesize
\begin{equation}
w_{k,d,j}=\frac{\sum_{n}\sum_{t=t_1}^{t_{n}}\gamma_{n,t,k}\mathds{1}_{Y_{n,t,d}}\mathds{1}_{Y_{n,t,d}=j}}{\sum_{n}\sum_{t=t_1}^{t_{n}}\sum_{j}\gamma_{n,t,k}\mathds{1}_{Y_{n,t,d}}\mathds{1}_{Y_{n,t,d}=j}}
\end{equation}}
{\footnotesize
\begin{equation}
\pi_{a}=\frac{\sum_{n}P(Z_{n,t_1}=a \vert Y_n;\pi',Q')}{\sum_{n}\sum_{k=1}^{K}P(Z_{n,t_1}=k \vert Y_n;\pi',Q')}
\end{equation}}
\noindent
The generator matrix $Q$ can be updated in each iteration using the closed-form solution:

{\scriptsize
\begin{equation}
Q_{ab}=\frac{\sum\limits_{\Delta}\sum\limits_{c,d \in [K]} \mathbb{E}[\mathcal{N}_{ab}(\Delta) \vert Z(\Delta)=d,Z(0)=c;Q']C_{c,d}(\Delta)}{\sum\limits_{\Delta}\sum\limits_{c,d \in [K]} \mathbb{E}[\mathcal{R}_{a}(\Delta) \vert Z(\Delta)=d,Z(0)=c;Q']C_{c,d}(\Delta)}
\end{equation}}
\noindent
The specific formulae for the involved terms are in \cite{Metzner2007,Wang2014}. The evaluation of posterior probabilities $\gamma_{n,t,k}$ and $P(Z_{n,t_{i-1}}=a,Z_{n,t_{i}}=b \vert {\bf Y};\pi',Q',w')$ is done using the forward-backward algorithm for computing the posterior probabilities in hidden Markov models; detailed derivations are in \cite{Bishop2007}.

%
%
%
%
%

\end{document}